\documentclass{IEEEoj}

\usepackage{cite}
\usepackage{amsmath,amssymb,amsfonts}
\usepackage{algorithmic}
\usepackage{graphicx,color}
\usepackage{textcomp}

\usepackage{algorithm}
\usepackage{booktabs}
\usepackage[switch]{lineno}

\def\BibTeX{{\rm B\kern-.05em{\sc i\kern-.025em b}\kern-.08em
    T\kern-.1667em\lower.7ex\hbox{E}\kern-.125emX}}
\AtBeginDocument{\definecolor{ojcolor}{cmyk}{0.93,0.59,0.15,0.02}}
\def\OJlogo{\vspace{-14pt}\includegraphics[height=10pt]{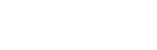}}

\begin{document}

\receiveddate{May, 2025. This work has been submitted to the IEEE for possible publication. Copyright may be transferred without notice, after which this version may no longer be accessible.}

\title{Reinforcement Learning-based Sequential Route Recommendation for System-Optimal Traffic Assignment}

\author{{\bf L}eizhen Wang\textsuperscript{1}, Peibo Duan\textsuperscript{1}*, \textit{Member, IEEE}, Cheng Lyu\textsuperscript{2} AND Zhenliang Ma\textsuperscript{3}*, \textit{Member, IEEE}}
\affil{Department of Data Science and Artificial Intelligence, Monash University, Melbourne, Australia}
\affil{Chair of Transportation Systems Engineering, Technical University of Munich, Munich, Germany}
\affil{Department of Civil and Architectural Engineering, KTH Royal Institute of Technology, Stockholm, Sweden}
\corresp{*CORRESPONDING AUTHOR: Peibo Duan, Zhenliang Ma}

\begin{abstract}
Modern navigation systems and shared mobility platforms increasingly rely on personalized route recommendations to improve individual travel experience and operational efficiency. However, a key question remains: can such sequential, personalized routing decisions collectively lead to system-optimal (SO) traffic assignment? This paper addresses this question by proposing a learning-based framework that reformulates the static SO traffic assignment problem as a single-agent deep reinforcement learning (RL) task. A central agent sequentially recommends routes to travelers as origin-destination (OD) demands arrive, to minimize total system travel time. To enhance learning efficiency and solution quality, we develop an MSA-guided deep Q-learning algorithm that integrates the iterative structure of traditional traffic assignment methods into the RL training process. The proposed approach is evaluated on both the Braess and Ortúzar–Willumsen (OW) networks. Results show that the RL agent converges to the theoretical SO solution in the Braess network and achieves only a 0.35\% deviation in the OW network. Further ablation studies demonstrate that the route action set's design significantly impacts convergence speed and final performance, with SO-informed route sets leading to faster learning and better outcomes. This work provides a theoretically grounded and practically relevant approach to bridging individual routing behavior with system-level efficiency through learning-based sequential assignment.
\end{abstract}



\maketitle

\section{INTRODUCTION}

\IEEEPARstart{T}{raffic} assignment is a classical and fundamental problem in the fields of transportation planning and traffic engineering, serving as the foundation for planning, designing, and managing transportation systems. The concept of traffic assignment revolves around the allocation of a predetermined origin-destination (OD) demand matrix onto a given urban road transportation network, based on pre-defined link performance functions. This optimization problem aims to predict and manage vehicular flow across the network, offering vital insights to guide urban planning and policy decisions \cite{patriksson2015traffic}.

Traditionally, traffic assignment problems are formulated to achieve either Wardrop's user equilibrium (UE) \cite{wardrop1952road,daskin1985urban} or system optimum (SO). The UE condition assumes that no driver can reduce their travel time by unilaterally changing routes, whereas the SO condition aims to minimize the total travel time for all drivers. These paradigms reflect a tension between decentralized individual choices and centralized social efficiency. Although SO provides the most efficient network use, it is often difficult to implement in practice due to uncertainties, lack of centralized control, and behavioral constraints \cite{morandi2024bridging}.

Extensive research has led to a wide range of algorithmic solutions for the UE problem, including link-based approaches such as Frank-Wolfe and path-based and bush-based formulations \cite{beckmann1956studies,bar2002origin}. While both UE and SO objectives are tractable within static traffic assignment frameworks, achieving SO in practice is considerably more difficult. This difficulty arises from the need for coordinated decisions across users, as well as practical issues such as limited user compliance and behavioral uncertainties \cite{ke2025real}. 

Reinforcement learning (RL) has recently gained attention for its potential to address complex decision-making problems in transportation. RL offers a flexible framework to learn routing policies by interacting with dynamic environments, without requiring complete system models. It has been applied in various domains, including traffic signal control \cite{chu2019multi,wang2023human}, autonomous driving \cite{kiran2021deep}, and energy management \cite{zou2016reinforcement}. In the context of traffic assignment, RL-based models have been explored to approximate UE solutions, by modeling OD-based route choices \cite{zhou2020reinforcement,stefanello2016using} and en-route link decisions \cite{grunitzki2014individual,mao2018reinforcement}. More recently, \cite{wang2024ai} proposed a large language model-based agent framework for UE assignment.

While these approaches have improved our understanding of decentralized routing behavior, the exploration of RL for achieving SO assignment remains limited. Compared to UE, the SO setting requires handling global optimization objectives and adapting to realistic driver behaviors. Notably, drivers may deviate from recommended routes that increase their perceived travel time, even if such recommendations are beneficial for the network. As highlighted in recent studies, this phenomenon of partial compliance can degrade the performance of system-optimal guidance, especially when users act independently \cite{yun2024navigating,bang2025route}.


Driven by the widespread use of real-time navigation systems, ride-hailing platforms, and shared mobility services, there is a growing need for intelligent route recommendation algorithms that operate in an online, sequential fashion. These systems typically assign routes to travelers as they appear, based on current traffic conditions and operational objectives. A fundamental and practical question thus arises: \textit{Can sequential, personalized route recommendations collectively yield a system-optimal traffic state?} Recent studies suggest that adaptive, learning-based route recommendations may help bridge the gap between user behavior and system-level efficiency \cite{bang2025route,yun2024navigating}. However, these efforts primarily focus on simulation-based evaluation of user behavior and policy learning, often lacking a rigorous comparison with the theoretical SO.

To address this challenge, this paper reformulates the static SO traffic assignment as an Markov Decision Process (MDP), where a central RL-based agent sequentially assigns routes as OD demands arrive. Using an analytical framework with known network structure and performance functions, we enable direct comparison with the theoretical SO solution and precise quantification of the optimality gap. We further develop a deep Q-learning algorithm guided by the Method of Successive Averages (MSA), integrating classic assignment insights with reinforcement learning. Our main contributions are as follows:

\begin{enumerate}
    \item We reformulate static SO traffic assignment as an MDP, where a centralized RL-based agent sequentially recommends routes.
    \item We propose an MSA-guided deep Q-learning algorithm that integrates classical traffic assignment principles to improve learning efficiency and policy convergence.
    \item We validate the method on both the Braess \cite{zhuang2022braess} and Ortúzar–Willumsen (OW) \cite{de2024modelling} networks, showing that it closely approximates the theoretical SO solution and highlighting the role of action space design in performance.
\end{enumerate}

The remainder of the paper is organized as follows: Section~\ref{sec:problem_formulation} defines the reinforcement learning-based problem formulation. Section~\ref{sec:method} details the proposed methodology. Section~\ref{sec:experiment} presents experimental validation and analysis. Section~\ref{sec:conclusion} concludes with a discussion of findings and future directions.

\section{PROBLEM FORMULATION}
\label{sec:problem_formulation}

We reformulate the static SO traffic assignment problem as an online sequential route recommendation problem, where routes are recommended one-by-one to individual travelers in real time.


The urban transportation network is modeled as a directed graph $G(V,E)$, where $V$ denotes the set of intersections (nodes), and $E$ denotes the set of road segments (edges). Each edge $e \in E$ is associated with a performance function $c(x_e)$, which maps the traffic flow $x_e$ on edge $e$ to the corresponding travel time.

Let $D = \{(o_i, d_i)\}_{i=1}^{N}$ represent a sequence of OD travel demands, where each $(o_i, d_i)$ denotes the origin and destination of the $i$-th traveler, and $N$ is the total number of demands. These travelers appear sequentially. The goal is to develop a decision-making agent that recommends a route from $o_i$ to $d_i$ for each traveler $i$ based on the current network state. The objective is to minimize the total travel time for all travelers, in alignment with the principle of SO traffic assignment.

This reformulation departs from classical traffic assignment by introducing a sequential decision process. Traditional static traffic assignment assumes that all OD demands occur simultaneously and are assigned collectively. In contrast, we model a more realistic setting in which travel requests arrive sequentially, and routing decisions must be made sequentially. Rather than computing a global static assignment, the central agent incrementally assigns routes to travelers to optimize the cumulative system performance.

\textbf{Remark.}
\begin{enumerate}
    \item We assume full compliance: all travelers follow the route recommended by the central recommendation agent. This allows us to isolate and evaluate the policy's ability to approach SO outcomes.
    \item Our formulation differs from \cite{shou2022multi}, which models each traveler as an individual agent in a multi-agent setting. Here, we consider a single centralized agent interacting within a static traffic assignment environment to learn an optimal routing policy.
\end{enumerate}



\section{Methodology}
\label{sec:method}



We propose a single-agent deep Q-learning-based method to model the SO traffic assignment in the routing game. It is modeled as an MDP process. The MDP is represented as $\langle S, A, P, R, \gamma \rangle$, where $S$ is the set of state representations, $A$ is the possible actions set, $P$ is the state transition matrix, $R$ is the reward function, and $\gamma$ is the discount factor for accumulated returns. An MDP proceeds as follows: the state transitions according to the state $s_t$ and the chosen action $a_t$ at time step $t$:

\begin{equation}
    P_{ss'} = p(s_{t+1} = s' \mid s_t = s, a_t = a): S \times A \rightarrow S
\end{equation}

The agent receives an associated reward $r$ at time step $t$:

\begin{equation}
    r_t = r(s_t, a_t): S \times A \rightarrow \mathbb{R}
\end{equation}

The route recommendation agent finds a policy that maximizes the expected return:

\begin{equation}
    G_t := \sum_{i=0}^{\infty} \gamma^i \cdot r_{t+i}
\end{equation}

where $\gamma$ is a number between 0 and 1 that controls the importance of immediate rewards compared with future rewards.

An action-value function (Q-function) is used to estimate the policy $\pi$ with the cumulative discounted sum of rewards:

\begin{equation}
    Q^{\pi}(s_t, a_t) = \mathbb{E}^{\pi} [ G_t \mid s_t, a_t ]
\end{equation}

The optimal Q-function comes from the optimal policy that maximizes the discounted expected return, which satisfies the Bellman optimality equation:

\begin{equation}
    Q^*(s_t, a_t) = \mathbb{E} \left[ r_t + \gamma \cdot \max_{a_{t+1}} Q^*(s_{t+1}, a_{t+1}) \right]
\end{equation}

\begin{figure}
    \centering
    \includegraphics[width=0.45\textwidth]{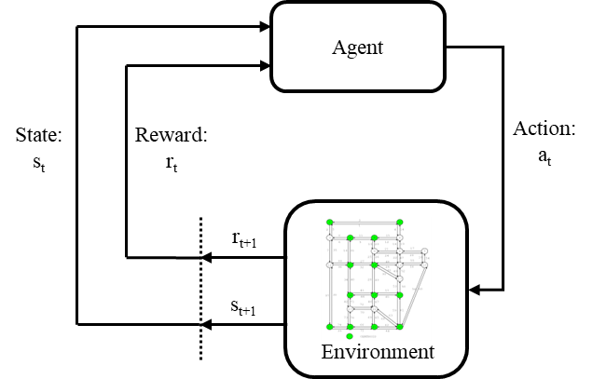}
    \caption{RL framework}
    \label{fig:rl_framework}
\end{figure}

Figure~\ref{fig:rl_framework} illustrates the RL-based framework. At each time step $t$, the agent extracts real-time traffic information $s_t$ from the road network environment. The agent recommends a route, or action, $a_t$ based on state $s_t$. The environment returns a reward $r_t$ as feedback to the agent and transitions to a new state $s_{t+1}$. A large amount of paired data $\langle s_t, a_t, r_t, s_{t+1} \rangle$ is generated by repeating the above process and stored in the replay buffer. Then, in deep Q-learning (DQN) \cite{mnih2015human}, the agent samples a batch of data randomly from the replay buffer to train neural networks to approximate the Q-function, $\hat{Q}(s_t, a_t; \theta) \approx Q^*(s_t, a_t)$. The recommendation policy is a mapping between the optimal action at each step (assigning a traveler to the network) and the currently observed environment state.

\subsection{Loss Function}
The loss function for updating neural networks is:

\begin{equation}
    J(w) = \mathbb{E} \left[ \left( \hat{Q}(s_t, a_t; \theta) - y_t \right)^2 \right],
\end{equation}

where

\begin{equation}
    y_t = r_t + \gamma \cdot \max_{a_{t+1}} \hat{Q}(s_{t+1}, a_{t+1}; \theta'),
\end{equation}

$\theta$ and $\theta'$ are neural network parameters from the current update step and the previous update step, respectively. Detailed information on updating process of double duel deep Q learning (DQN) can refer to \cite{mnih2015human,van2016deep,wang2016dueling}.

\subsection{Environment}
Let $\mathcal{T} = \{1, \dots, t, \dots, t_{max}\}$ denote the time of the network, which is when the $t$-th traveler requests a route recommendation, and the system is currently at time $t=1$. For any time $t$, we use the random vector $x_e^t$ to represent the traffic volume on road segments $e$ at time $t$. 

We use the link performance function $c_e (x_e^t)$ to calculate its travel time. For example, the commonly used non-linear performance function developed by the Bureau of Public Roads (BPR):

\begin{equation}
    \text{tt}_t^e = t_0^e \left( 1 + 0.15 \left( \frac{x_e^t}{w^e} \right)^4 \right),
\end{equation}

where $\text{tt}_t^e$ is the travel time of drivers on link $e$ when they enter the link at time $t$ given the assigned traffic volume $x_e^t$; the parameter $t_0^e$ is the free flow travel time of link $e$, and $w^e$ is its capacity.

\subsection{Action Definition}
The action space defines the feasible route set for an OD pair. An agent selects an route from the action set $A= \{ p_1, \dots, p_k, \dots, p_{|A|} \}$ where $p_k$ is the $k$-th predefined route choice, and $|A|$ is the number of route choices.

\subsection{Reward Function}
The reward is the target for the RL agent to maximize. We propose a reward function based on the marginal travel time of each traveler, which is the change in the total system travel time caused by a single traveler entering a route at a certain time interval. The marginal travel time $\text{tt}_{\text{marginal}}^p$ of a traveler on route $p$ is defined as the sum of travel time of the entering traveler and the additional delay of all travelers experienced on that route $p$:

\begin{equation}
    \text{tt}_{\text{marginal}}^p = -\sum_{e \in p} \left( c_e(v_e) + v_e \frac{dc_e(v_e)}{dv_e} \right),
\end{equation}

where $v_e$ is the assigned traffic volume of link $e$, $c_e(v_e)$ is the travel time of link $a$, and $v_e \frac{dc_e(v_e)}{dv_e}$ is the additional travel time burden that the newly assigned traveler inflicts on each of the other travelers.

\subsection{State Representation}
The agent makes decisions and estimates expected returns based on state inputs. In order to provide the RL model with effective information, we provide two kinds of information (edge-level and OD-level information):

\begin{equation}
    S_{\text{edge}}^t = \{ \text{tt}_t^e, v_t^e, \text{tt}_{\text{marginal}}^e \}_{e=1}^{|E|},
\end{equation}

\begin{equation}
    S_{\text{od}}^t = \{ o_{\text{od}}, \text{tt}_{\text{marginal}}^{p_k} \}_{k=1}^{|A|},
\end{equation}

where $o_{\text{od}}$ is a one-hot vector representing the origin and destination of the traveler.

\subsection{MSA-Guided RL Algorithm}

DQN employs the $\epsilon$-greedy strategy to strike a balance between exploring new solution directions and exploiting learned policies, which is inefficient when dealing with large solution spaces. In this study, we integrate the traditional MSA used in traffic assignment to guide the learning process of the RL agent. Algorithm \ref{alg:msa_rl} shows the pseudo code of the MSA-guided RL algorithm. The main difference with DQN is that the action is selected based on Algorithm \ref{alg:msa_selection} in each training step, by sampling actions from the assignment distribution iterated by the MSA method. 

The ``all-or-nothing" algorithm \cite{boyles2020transportation} is commonly used in traffic assignment, where each trip (OD pair) is assigned to the shortest path between the origin and destination, without considering any other alternative routes. After each episode, the assignment distribution $M$, which represents the proportion of demand allocated to different optional routes for each OD pair, is optimized based on the flow assigned in the current episode. Note that the optional routes are also added incrementally in line with the MSA approach. The utilization of the MSA method effectively guides the RL agent to explore towards SO traffic assignment and update the action (route) set.

\begin{algorithm}
\caption{MSA-Guided RL Algorithm}
\label{alg:msa_rl}
\begin{algorithmic}[1]
\STATE \textbf{Initialize:} Replay buffer $D$, action-value function $Q$, target action-value function $Q^-$, learning rate $\alpha$, discount factor $\gamma$, a small number $\epsilon$, road segments flow $V$, assignment distribution $M$
\FOR{episode $i = 1$ to $M$}
    \STATE Initialize state $S$
    \WHILE{$S$ is not terminal}
        \STATE $A \leftarrow \text{MSA\_Guided\_Selection}(Q, S, \epsilon, M)$ 
        \STATE Take action $A$, then update road segments flow $V$, observe reward $R$ and next state $S'$
        \STATE $Q(S,A) \leftarrow Q(S,A) + \alpha \cdot [R + \gamma \cdot \max\limits_{a} Q^- (S', a) - Q(S,A)]$
    \ENDWHILE
    \STATE $M^* \leftarrow \text{all\_or\_nothing}(V)$
    \STATE $M \leftarrow (1 - \frac{1}{i}) \cdot M + \frac{1}{i} M^*$
\ENDFOR
\end{algorithmic}
\end{algorithm}

\begin{algorithm}
\caption{MSA-Guided Selection}
\label{alg:msa_selection}
\begin{algorithmic}[1]
\STATE $n \leftarrow$ uniform random number between 0 and 1
\IF{$n < \epsilon$}
    \STATE $A \leftarrow$ sample from $M$
\ELSE
    \STATE $A \leftarrow \arg\max Q(S)$
\ENDIF
\STATE \textbf{Return} $A$
\end{algorithmic}
\end{algorithm}

\section{Case Study}
\label{sec:experiment}

In this section, we demonstrate the effectiveness of RL in approximating the SO static traffic assignment. In Example 1, we apply DQN to a simple Braess paradox network. Example 2 scales this application to a relatively more complex OW Network \cite{de2024modelling}, incorporating a varying action set to elucidate the impact of action set modifications on the proposed algorithm's performance.

\subsection{Example 1: Braess Paradox}
\begin{figure}
    \centering
    \includegraphics[width=0.18\textwidth]{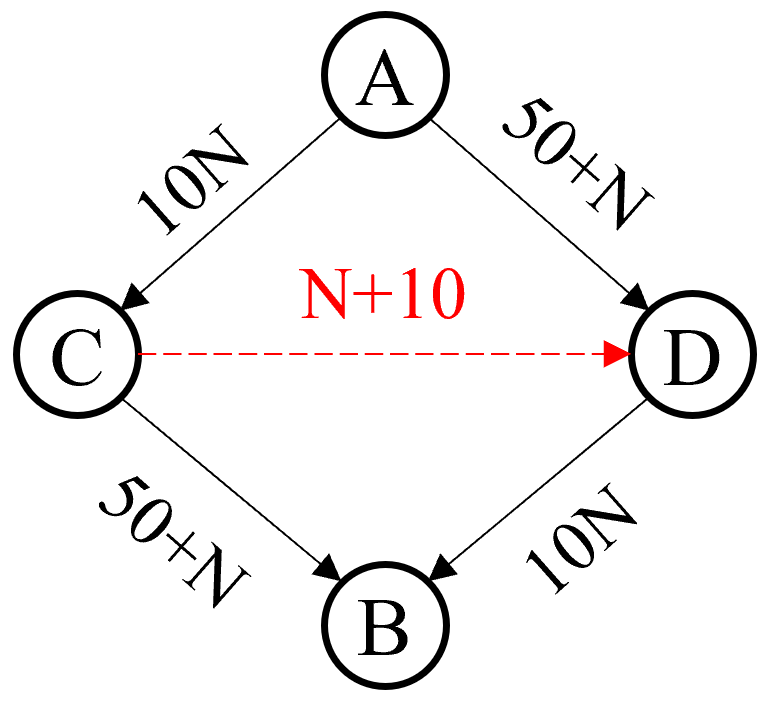}
    \caption{Network of the Braess Paradox}
    \label{fig:braess_net}
\end{figure}

Figure~\ref{fig:braess_net} illustrates the configuration of the Braess paradox network \cite{zhuang2022braess}, a diamond-shaped network with six vehicles traveling from $A$ to $B$. Each road segment's performance function is indicated next to the edge, where $N$ represents the number of vehicles utilizing that segment. Initially, the network provides three routes from $A$ to $B$, namely ACB and ADB. UE is achieved when three vehicles opt for route ACB, and the other three choose ADB, each route accruing a cost of 83 minutes. 

Interestingly, the introduction of a new road segment (depicted as the red dashed line in Figure~\ref{fig:braess_net}) connecting $C$ and $D$ alters the user equilibrium: two users each choose ACB and ADB, while the remaining two select ACDB, resulting in a route cost of 92 minutes, which exceeds the cost prior to the addition of the road. This paradox and the simulation in \cite{zhuang2022braess} highlight that increasing road capacity within a network could paradoxically increase travel time and often leads users to settle for sub-optimal network usage, a phenomenon termed the ``price of anarchy" \cite{morandi2024bridging}. It underscores the counterintuitive nature and complex dynamics of traffic networks, emphasizing the need for a more sophisticated strategy for enhancing network efficiency.

We first utilize this scenario to verify that our proposed method can converge to the SO traffic assignment by sequentially recommending routes to six travelers, even with the inclusion of the new route (ACDB). 

\begin{figure}
    \centering
    \includegraphics[width=0.48\textwidth]{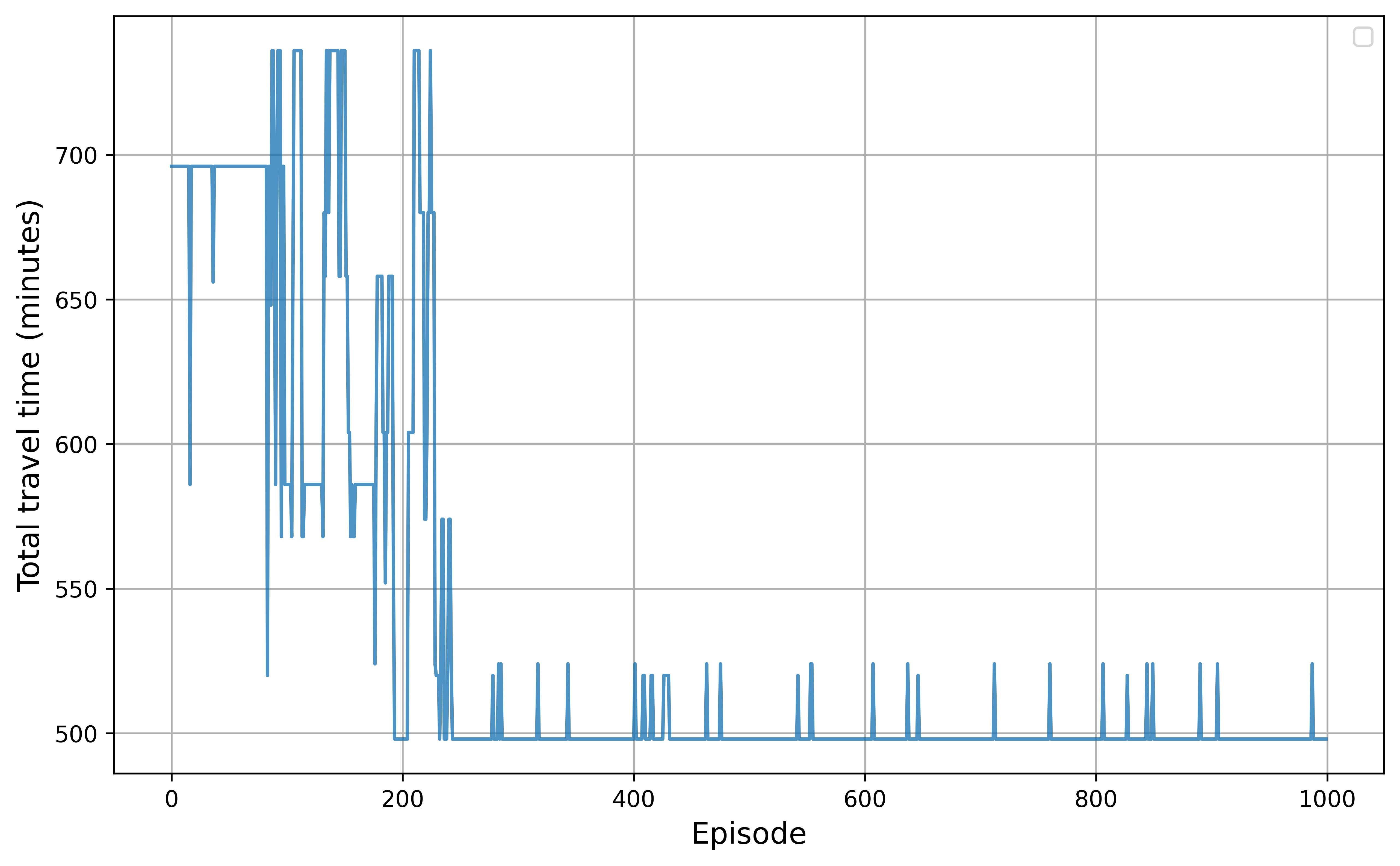}
    \caption{The training curve of the RL assignment model}
    \label{fig:braess_training_curve}
\end{figure}

Figure \ref{fig:braess_training_curve} depicts the training progression of the RL agent. After approximately 240 episodes, the total travel time (objective value) fundamentally stabilizes at 498  minutes (with a cost of 83 minutes for each traveler), achieving the SO traffic assignment. In this configuration, three travelers are recommended to ACB, and the remaining three to ADB. Notably, no traveler utilizes the route ACDB. This suggests that the RL agent can effectively converge to the SO by sequentially recommending routes for each traveler and interacting with the traffic environment.

\subsection{Example 2: OW Network}
In this example, we apply the proposed RL-based algorithm to the larger OW network, comprising 13 nodes, 48 links, and 1700 travelers—each operating a distinct vehicle \cite{de2024modelling}. The experiments are conducted under different action set configurations, where each action set (i.e., route choice set) is generated using the \( k \)-shortest paths algorithm \cite{yen1971finding} or traditional traffic assignment method.

\subsubsection{Experiment Settings}
\begin{figure}
    \centering
    \includegraphics[width=0.4\textwidth]{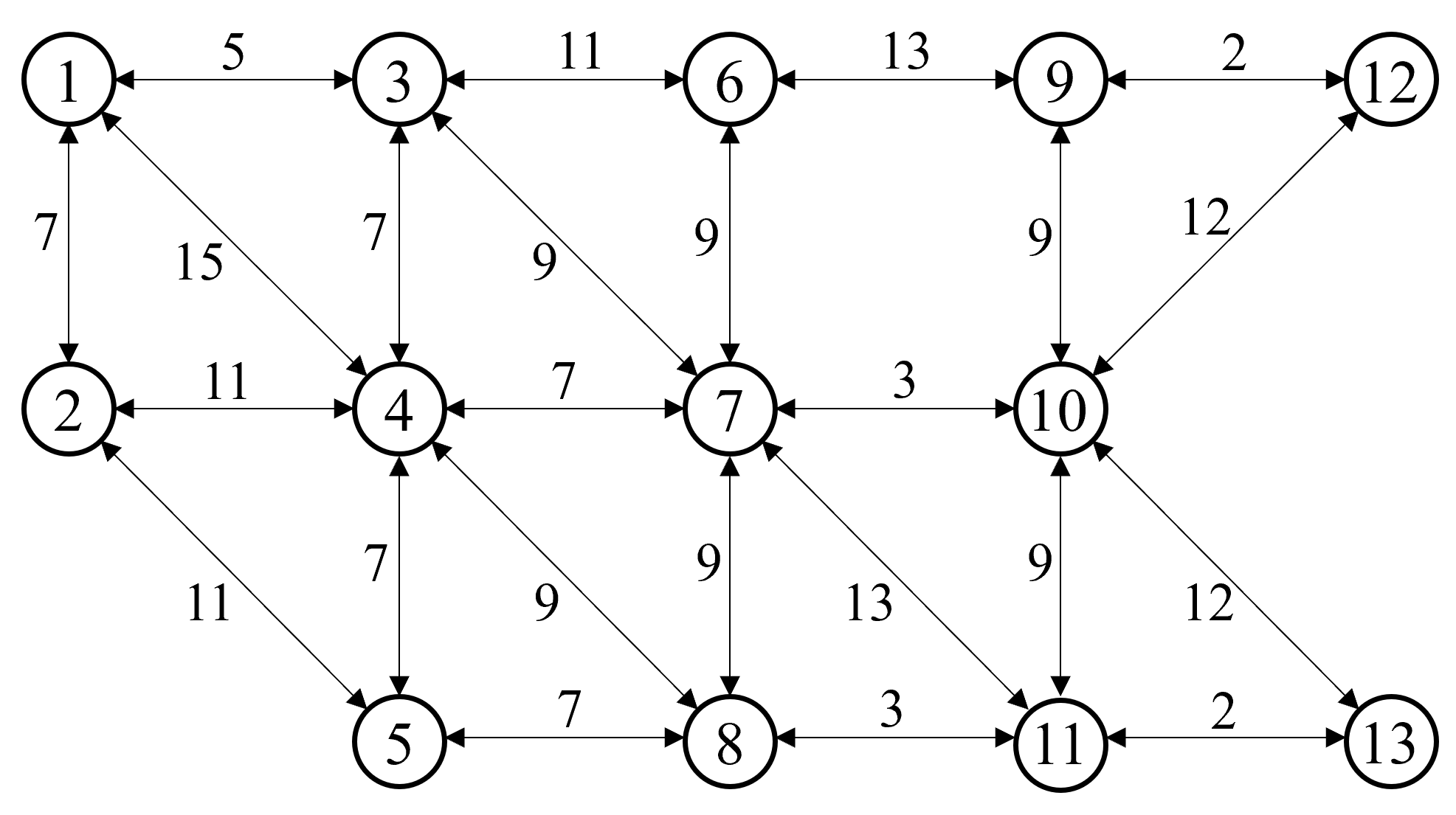}
    \caption{OW network}
    \label{fig:ow_network}
\end{figure}

Figure~\ref{fig:ow_network} shows the OW road network that connects two residential areas, Node 1 and Node 2, with two large shopping centers, Node 12 and Node 13. The travel times between these points are measured in minutes, and all links are two-ways. Table \ref{tab:od_demand} shows the OD demand of the OW network.

\begin{table}
    \centering
    \caption{OD demand of OW network}
    \label{tab:od_demand}
    \begin{tabular}{ccc}
        \toprule
        Origin & Destination & Demand \\
        \midrule
        1 & 12 & 600 \\
        1 & 13 & 400 \\
        2 & 12 & 300 \\
        2 & 13 & 400 \\
        \bottomrule
    \end{tabular}
\end{table}

The hyperparameters for training the RL-based assignment model were selected based on grid search. The key configurations are listed in Table~\ref{tab:rl_hyperparameters}.

\begin{table}[ht]
    \centering
    \caption{Hyperparameter settings}
    \small
    \begin{tabular}{cc}
        \toprule
        \textbf{Hyperparameter} & \textbf{Value} \\ \midrule
        Hidden Layers (Units) & 2 (512, 256) \\
        Discount Factor ($\gamma$) & 0.95 \\
        Mini-Batch Size & 128 \\
        Learning Rate & $2 \times 10^{-5}$ \\
        \bottomrule
    \end{tabular}%
    
    \label{tab:rl_hyperparameters}
\end{table}

\subsubsection{Compared Methods}

The performance of the proposed method is compared with the following configurations:

\begin{itemize}
    \item \textbf{RL with K-shortest paths action set (RL-K-SP):} RL model using an action set composed of $K$-shortest paths for each OD pair.
    
    \item \textbf{RL with system-optimum route paths (RL-SO):} RL model using action sets derived from the analytical SO solution, i.e., the route set generated from a traditional traffic assignment algorithm.
    
    \item \textbf{MSA-guided RL:} RL model trained with a strategy guided by the MSA to explore and update the assignment policy.
\end{itemize}




\subsubsection{Experiment Results and Analysis}

Table~\ref{tab:traditional_results} presents the total travel time and relative gap optimized by traditional methods, including the MSA and Frank-Wolfe (FW), both targeting UE and SO solutions on the OW network over 10{,}000 iterations. The MSA method yields two benchmark solutions that are considered ``close enough''—that is, solutions with a relative gap less than or equal to $10^{-4}$, which is typically regarded as sufficiently close to optimal \cite{boyles2020transportation}. These benchmark results serve as baselines for evaluating the performance of the RL-based methods.

\begin{table}
    \centering
    \caption{Traditional method results}
    \label{tab:traditional_results}
    \begin{tabular}{cccc}
        \toprule
        Method & Total Travel Time (min) & Iterations & Relative Gap \\
        \midrule
        UE-MSA & 57052.1 & 10000 & 0.00001 \\
        SO-MSA & 54809.8 & 10000 & 0.00006 \\
        UE-FW & 57055.9 & 10000 & 0.00017 \\
        SO-FW & 54808.5 & 10000 & 0.00003 \\
        \bottomrule
    \end{tabular}
\end{table}

\begin{figure}
    \centering
    \includegraphics[width=0.45\textwidth]{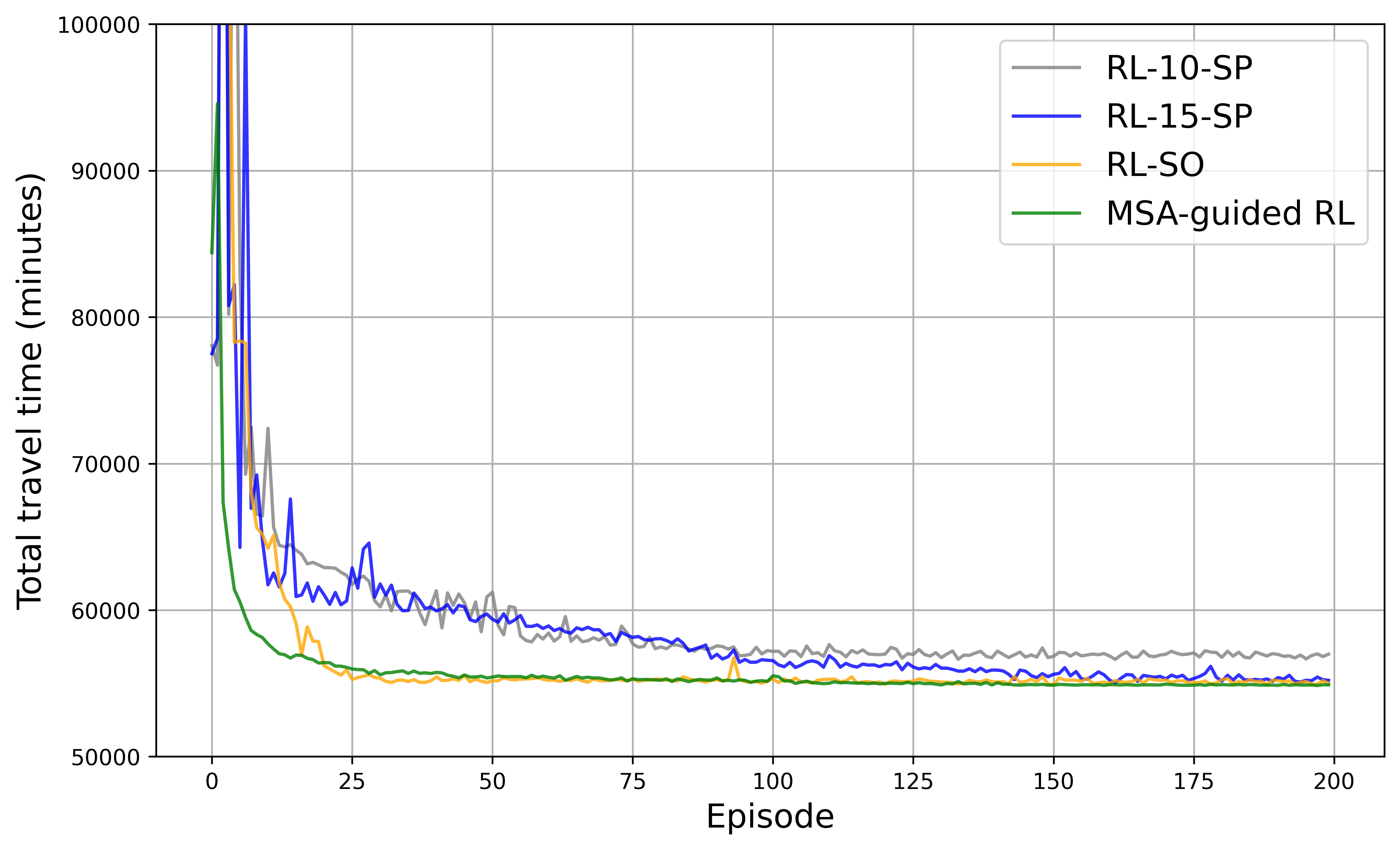}
    \caption{The training curve of all RL-based assignment models}
    \label{fig:OW_training_curve}
\end{figure}

Figure~\ref{fig:OW_training_curve} illustrates the training progression of all RL-based agents. Both RL-SO and MSA-guided RL exhibit fast convergence and close approximation to the SO solution. In contrast, agents with larger action spaces, such as RL-15-SP, show slower convergence but improved final performance due to their greater exploratory capacity compared to RL-10-SP. This highlights the impact of action space size on both solution quality and learning dynamics.

Table~\ref{tab:rl_results} shows the test results of all RL-based methods. The metric ``(UE$-x$)/UE" measures the improvement over UE-MSA, where higher values indicate better performance. Conversely, ``($x$-SO)/SO" measures the deviation from the optimal SO-MSA solution, with smaller values being more desirable. Most RL-based methods achieve acceptable performance, except RL-10-SP, which reduces travel time by only 0.58\% compared to UE-MSA. This may be due to the 10-shortest-path action set omitting some important routes included in the analytical solution.

\begin{table}
    \centering
    \caption{RL-based method test results}
    \label{tab:rl_results}
    \begin{tabular}{cccc}
        \toprule
        Method & Total Travel Time (min) & (UE$-x$)/UE & ($x$-SO)/SO \\
        \midrule
        UE-MSA & 57052.1 & -- & -- \\
        SO-MSA & 54809.8 & 3.93\% & -- \\
        MSA-guided RL & 55000 & 3.60\% & 0.35\% \\
        RL-SO & 54950 & 3.68\% & 0.26\% \\
        RL-10-SP & 56720 & 0.58\% & 3.49\% \\
        RL-15-SP & 55220 & 3.21\% & 0.75\% \\
        \bottomrule
    \end{tabular}
    
    \vspace{2mm}
\end{table}

Additionally, all RL-based methods slightly underperform the SO-MSA benchmark. Specifically, limiting the action set to 10 shortest paths results in a 3.49\% deviation from the SO. This deviation reduces to 0.75\% when expanding the action set to 15 shortest paths. When the action set is directly generated from the SO-MSA solution, the RL model's performance improves significantly, achieving a deviation of only 0.26\% and faster convergence.

These results underscore the critical role of the action set in shaping RL agent performance, affecting both convergence speed and solution quality. Our method, which jointly learns the recommendation policy and incrementally updates the action set, achieves a comparable deviation of 0.35\% from SO-MSA. Although the MSA-guided RL is not the top-performing variant, it remains effective when the SO route set is unavailable—an assumption that aligns more closely with practical applications.

\section{CONCLUSION}
\label{sec:conclusion}

This paper reformulates the SO traffic assignment problem as an MDP and presents a novel reinforcement learning-based algorithm to solve it. Unlike many prior works that rely on simulation-based evaluation, our framework is fully analytical—allowing direct comparison with the theoretical SO benchmark and enabling precise quantification of the learned policy's optimality gap.

Experimental results demonstrate the algorithm's capability to identify the theoretical SO solution on the Baress paradox network. It can also approximate the analytical SO solution within a 0.35\% margin for a more complex OW network. This highlights the potential of reinforcement learning to not only replicate, but also systematically approach optimal routing outcomes when embedded in a theoretically grounded assignment framework. In addition, we find that the design of the action space significantly affects the convergence speed and final performance.

The primary objective of this study is not to design the most advanced RL architecture, but to explore the potential of reformulating static SO traffic assignment as a sequential learning problem with analytical validation. This perspective opens new directions for integrating learning-based methods into classical transportation optimization tasks. Future work will extend this framework to account for partial user compliance, stochastic demand, and dynamic traffic conditions, thereby improving its realism and practical applicability.

\section*{ACKNOWLEDGMENT}
The work was supported by start-up funds with No. MSRI8001004 and No. MSRI9002005 at Monash University and TRENOP fund at KTH Royal Institute of Technology, Sweden.

\section*{AUTHOR CONTRIBUTIONS}
The authors confirm contribution to the paper as follows: study conception and design: Z Ma, L Wang, C Lyu; methodology: P Duan, L Wang, Z Ma, C Lyu; data collection: L Wang, C Lyu; analysis and interpretation of results: L Wang, C Lyu, Z Ma, P Duan; draft manuscript preparation: L Wang, Z Ma, P Duan. manuscript revision: Z Ma, P Duan, C Lyu. All authors reviewed the results and approved the final version of the manuscript.

\bibliographystyle{IEEEtran}
\bibliography{references}

\end{document}